\documentclass[12pt]{article}
\usepackage{amsmath}
\usepackage{graphicx}
\usepackage{enumerate}
\usepackage{natbib}
\usepackage{url} 
\usepackage{longtable}
\usepackage{adjustbox}
\usepackage{diagbox}
\usepackage{makecell}
\usepackage[figuresright]{rotating}
\usepackage{multicol,multirow}
\usepackage{amssymb}

\newcommand{\blind}{1}

\DeclareUnicodeCharacter{2061}{}

\begin{document}

\bibliographystyle{agsm}

\def\spacingset#1{\renewcommand{\baselinestretch}%
{#1}\small\normalsize} \spacingset{1}


\if1\blind
{
  \title{\bf Unveiling LLM Evaluation Focused on Metrics: Challenges and Solutions\footnote{This research was funded by supported by National Key R\&D Program of China (No. 2021YFF0901400)}}
  \author{Taojun Hu\hspace{.2cm}\\
    Department of Biostatistics, Peking University\\
    and \\
    Xiao-Hua Zhou\thanks{Correspondence author, azhou@math.pku.edu.cn} \\
    Department of Biostatistics, Peking University\\
    Chongqing Big Data Research Institute, Peking University\\
    Beijing International Center for  Mathematical Research, Peking University}
  \maketitle
} \fi

\if0\blind
{
  \bigskip
  \bigskip
  \bigskip
  \begin{center}
    {\LARGE\bf Unveiling LLM Evaluation Focused on Metrics: Challenges and Solutions}
\end{center}
  \medskip
} \fi

\bigskip
\begin{abstract}
Natural Language Processing (NLP) is witnessing a remarkable breakthrough driven by the success of Large Language Models (LLMs). LLMs have gained significant attention across academia and industry for their versatile applications in text generation, question answering, and text summarization. As the landscape of NLP evolves with an increasing number of domain-specific LLMs employing diverse techniques and trained on various corpus, evaluating performance of these models becomes paramount. 
To quantify the performance, it's crucial to have a comprehensive grasp of existing metrics.
Among the evaluation, metrics which quantifying the performance of LLMs play a pivotal role. This paper offers a comprehensive exploration of LLM evaluation from a metrics perspective, providing insights into the selection and interpretation of metrics currently in use. Our main goal is to elucidate their mathematical formulations and statistical interpretations. We shed light on the application of these metrics using recent Biomedical LLMs. Additionally, we offer a succinct comparison of these metrics, aiding researchers in selecting appropriate metrics for diverse tasks. The overarching goal is to furnish researchers with a pragmatic guide for effective LLM evaluation and metric selection, thereby advancing the understanding and application of these large language models.
\end{abstract}

\noindent%
\vfill

\newpage
\spacingset{1.9} 
\section{Introduction}
\label{sec:intro}

ChatGPT, also namely GPT3.5, \citep{brown2020language,wang2022self} has demonstrated its remarkable ability to generate coherent sequences of words and engage in conversational interactions. As the popularity of ChatGPT surged, quantities of large language models (LLMs) emerged rapidly. Researchers not only explore distinct model architectures and propose various fine-tuning methods to enhance LLM capabilities but also investigate ways to tailor LLMs to specific domains.

LLMs present a significant opportunity for tasks such as generating scientific texts, answering questions, and extracting core information from articles. For instance, amidst the daily influx of over 3000 new articles in peer-reviewed journals \citep{sayers2023database}, LLMs can swiftly extract key insights, aiding readers in navigating through vast amounts of medical literature. Furthermore, LLMs can potentially analyze symptoms described by patients to suggest diagnoses and treatment options, thus alleviating physicians' workload and improving patient care, which is a useful application in medicine. These applications underscore two primary functionalities of LLMs: information retrieval and text generation. However, LLMs can also contribute to various other aspects of medical research and applications. Due to these advantages, today's LLMs are attracting increasing attention across multiple fields. With the surge of novel LLMs, reviewing and evaluating existing LLMs is of great importance. 
Among evaluations, the intrinsic statistical interpretations are frequently neglected in current research and reviews.

The proliferation of LLMs has prompted the emergence of reviews aimed at providing insights into their development and potential applications. For instance, \cite{naveed2023comprehensive} reviewed LLMs from 2019, starting with T5, up to the latest releases in 2023, offering comprehensive references and comparisons. \cite{wang2023pre} highlighted the promising applications of LLMs in addressing biomedical questions, while \cite{chen2023large} focused on evaluating biomedical LLMs with respect to benchmarks and summarization capabilities. While existing reviews generally discuss LLM structures and applications, evaluating LLMs is crucial for guiding their development and deployment but understudied. Evaluation encompasses various aspects, including downstream tasks, criteria, benchmark datasets, and metrics. Although \cite{chang2023survey} surveyed LLM evaluation, comprehensive summarizing the metrics remains scarce. This work aims to fill this gap by providing a survey of contemporary LLM evaluation metrics, along with mathematical formulations and statistical explanations and practical guidance for implementation using open-source libraries. Our paper shed lights on the bridge between the existing LLM evaluations and statistics by exploring the statistical interpretations of the metrics. Additionally, we showcase how these metrics have been utilized in conjunction with published biomedical LLMs through illustrative examples. Our main contributions are four-folds:
\begin{itemize}
    \item Summarizing and categorizing the metrics for LLM evaluation into three distinct types;
    \item Providing mathematical formulations and statistical interpretations for each metric, along with a comparative analysis to serve as a guide for LLM researchers;
    \item Identifying repositories containing the discussed metrics;
    \item Showcasing how these metrics and baseline datasets are applied in the evaluation of recently developed biomedical LLMs, facilitating further studies to keep alignment with previous models.
\end{itemize}

The article is structured as follows: Section 2 offers a concise overview of LLM evaluation criteria. Section 3 details the most utilized metrics in LLM evaluations, including their mathematical expressions and statistical interpretations, alongside a directory of repositories for implementing these metrics. Section 4 showcases the application of these metrics using biomedical LLMs as case studies, including the baseline datasets employed for evaluation. Section 5 discusses the pros and cons of the existing widely used metrics and stresses two major common problems that are often ignored: the imperfect gold standard issue and absent of statistical inference method. The paper concludes with Section 6, summarizing our findings and the limitations for this paper.

\section{Evaluations for LLMs}
\label{sec:back}

We provide a brief overview of evaluation criteria for LLMs, which, although not the main focus of this paper, are essential for understanding critical development aspects of LLMs. With NLP's long history, models have been developed for specific tasks, either supervised or unsupervised. Accuracy in generating a desired answer is a predominant focus in LLM evaluations. However, issues such as overfitting and ignoring detrimental aspects like unfairness or hallucination render accuracy an imperfect evaluation metric. Hallucination, gaining researchers' attention recently, involves generating false or misleading information by well-trained LLMs, often due to biases in training data, leading to overconfident and inaccurate outputs. This overconfidence is closely linked to an overreliance on accuracy-oriented training. To address these issues, it's crucial to understand the key considerations and criteria in LLM evaluation. We discuss these evaluation criteria from various perspectives in this section.

\paragraph{Accuracy} \citep{bengio2000neural,morin2005hierarchical} measures how well the LLM produces desired results, a primary performance concern. Typically, gold standard answers are used, or specialists are employed to assess LLM performance. High accuracy ensures optimal quality and unbiased predictions, aligning LLMs with user needs and instructions, making it a fundamental requirement.

\paragraph{Ethicality} \citep{weidinger2021ethical,ganguli2022red,jobin2019global} involves a broad range of concerns, including privacy protection, misinformation reduction, fairness, and transparency. Given LLMs are trained on large datasets that may include sensitive information or deceptive content, ethicality mandates LLMs to produce legal, safe, and ethical outputs. Ongoing research, transparent practices, and thoughtful policy-making are essential for ensuring LLMs' positive societal contributions.

\paragraph{Fairness} \citep{bolukbasi2016man,dixon2018measuring,hovy2021importance,selbst2019fairness}, a critical aspect of ethicality with significant social implications, demands equal treatment from LLM outputs, regardless of individual or group demographics. It requires bias mitigation to prevent discriminatory decisions, highlighted by the need for fairness in pronoun prediction in sentences like "\_\_ is a nurse,". A fair system of LLMs is supposed to put no preference on "he" or "she" that suggests the sex directly. The criterion of fairness promotes unbiased system responses.

\paragraph{Generalization} \citep{raffel2020exploring,hupkes2023taxonomy,lazaridou2021mind} indicates an LLM's ability to adapt to unseen data, crucial for responding to diverse queries and understanding text generation mechanisms. Techniques like regularization and diverse dataset training enhance generalization, which is key for LLMs to comprehend language and context broadly.

\paragraph{Robustness} \citep{wang2021measure,goel2021robustness,goyal2023survey} describes an LLM's resilience to errors, manipulations, or adversarial attacks, aiming for trustworthy and consistent outputs. Addressing this involves varied dataset training and adversarial methods to ensure performance stability and real-world application trustworthiness.

\paragraph{Reasoning} \citep{valmeekam2022large,jin2023cladder}, or an LLM's ability to logically infer or deduce information, is essential for applying learned knowledge to new contexts. This capability, requiring further research, underscores the need for LLMs to exhibit logical and causal reasoning.

Evaluating LLMs comprehensively requires examining not only accuracy but also ethicality, fairness, generalization, robustness, and reasoning. Each aspect is vital for creating intelligent LLMs that benefit society and users. Employing benchmark datasets is a common solution for comprehensive evaluation, with metric design playing a crucial role in assessing LLM capabilities, which we explore further in the next section.

\section{Metrics}
\label{sec:metrics}

The metrics can be broadly categorized into three types. The first type, which assesses the ability to accurately classify texts into at least two labels, is the most prevalent. We refer to these as the \textbf{Multiple-Classification (MC) metrics}. They can be applied to various tasks with developed benchmark datasets. 
The second type, known as the \textbf{Token-Similarity (TS) metrics}, evaluates how well the generated texts align with the expected texts.
Lastly, a metric specifically tailored for the Question-Answering task is the \textbf{Question-Answering (QA) metrics}. 
In the following sections, we will provide a detailed illustration of each metric along with its mathematical formulation. Meanwhile, we also explain for these metrics from a statistical perspective. 

\subsection{Multiple-Classification metrics}

The Multiple-Classification (MC) metrics assess how effectively the LLM classifies texts into multiple groups, with each group serving as a label. These metrics encompass Accuracy (Acc), Recall (R), Precision (P), and F1 scores \citep{goutte2005probabilistic}, particularly in two-label classification scenarios. In multi-label classification, the micro-F1 and macro-F1 metrics \citep{ghamrawi2005collective} are commonly employed.
To illustrate, let's consider a two-label problem, distinguishing between "positive" and "negative" cases. Benchmark datasets provide the gold standard label for each case, while the LLM performs classification and generates predictions. This process results in a 2×2 chart (Table~\ref{tab:tab1}), with cells representing True Positive (TP), False Positive (FP), False Negative (FN), and True Negative (TN) instances. Thus, Accuracy (Acc) can be expressed as:

\begin{equation}
    \mathrm{Acc} = \frac{TP + TN}{TP + FN + FP + TN}
\end{equation}

\begin{table}
    \centering
    \begin{tabular}{ccc}
    \hline
        True/Prediction &  Positive & Negative \\
        \hline
         Positive & \(y_{11}(TP)\) & \(y_{10}(FN)\) \\
         Negative & \(y_{01}(FN)\) & \(y_{00}(TN)\) \\
         \hline
    \end{tabular}
    \caption{The contingency table for a two-label classification problem}
    \label{tab:tab1}
\end{table}

The Recall, also known as Sensitivity or True Positive Rate, signifies the proportion of positive detections among the actual "positive" instances. It gauges an LLM's ability to identify positive instances. The formula for the Recall is given by:

\begin{equation}
    \mathrm{Recall} = \frac{\mathrm{TP}}{\mathrm{TP} + \mathrm{FN}}
\end{equation}

Precision, also referred to the Positive Predictive Value, indicates the proportion of actual positive instances among all instances identified as positive. It measures an LLM's precision in filtering out negative instances falsely labeled as positive. The Precision formula is:

\begin{equation}
    \mathrm{Precision} = \frac{\mathrm{TP}}{\mathrm{TP} + \mathrm{FP}}
\end{equation}

However, high Recall often corresponds to low Precision, and vice versa. The F1 score balances these two metrics using a simple harmonic mean operation:

\begin{equation}
    \mathrm{F1} = \frac{2 \times \mathrm{Recall} \times \mathrm{Precision}}{\mathrm{Recall} + \mathrm{Precision}}
\end{equation}

The F1 score ranges from 0 to 1, with F1\( = 1\) indicating both perfect Recall and perfect Precision.

From a statistical perspective, the aforementioned metrics carry inherent probabilistic interpretations. Before delving into their statistical interpretations, it's important to briefly define the sample and population of these metrics, an aspect often overlooked by researchers. The population for LLMs refers to the global corpus on which the models are evaluated, while the sample comprises the test datasets. In most cases, the test datasets are assumed to be randomly sampled from the global corpus, allowing the metrics computed on them to represent the model's performance on the global corpus.

Let's consider the classification outcomes provided by the LLM denoted as $X$, and the true labels as $Y$, both binary variables within $\{0, 1\}$. The Accuracy metric represents $P(X=Y) = P(X=1, Y=1) + P(X=0, Y=0)$, while Recall denotes $P(X=1\mid Y=1)$ and Precision denotes $P(Y=1\mid X=1)$. The F1-score means the harmonic mean of the previous two metric, that is 
$\frac{2}{1/P(X=1\mid Y=1) + 1/P(Y=1\mid X=1)}$.
The Receiver Operating Characteristic (ROC) curve, along with the Area Under the Curve (AUC), offers a comprehensive assessment of a classifier or diagnostic tool. It illustrates all possible trade-offs between sensitivity and specificity across various decision cut-offs. Similarly, the Precision-Recall (PR) curve showcases the interplay between Precision and Recall under different cut-offs, with Recall on the x-axis and Precision on the y-axis. The area under the Precision-Recall curve (PRAUC) quantifies the classifier's performance. Studies by \cite{saito2015precision} have demonstrated that PRAUC can provide more informative insights with imbalanced datasets where the negatives far outnumber the positives. However, to estimate AUC/PRAUC, a continuous biomarker for classification is necessary. Consequently, AUC/PRAUC evaluation may not be directly applicable to LLMs lacking such continuous biomarkers. Both the AUC and PRAUC offer comprehensive evaluations for LLMs without relying on specific cutoff values. Despite their potential advantages, many studies on LLMs have not reported AUC/PRAUC results, even when continuous biomarkers are available. This omission limits the comprehensiveness of evaluating LLM performance.

In multi-label cases, where there are more than two labels, the accuracy and two variants of the F1 score are widely used: micro-F1 and macro-F1. We illustrate the notations first. Suppose there are \(L\) labels marked from \(1\) to \(L\), then the number of instances that are classified as label \(i\) by the LLM but belong to the label \(j\) is noted by \(y_{{ij}}\). We have an \(L \times L\) matrix with elements \(\left( y_{{ij}} \right)_{1 \leq i \leq L,\ 1 \leq j \leq L}\). Then the simplest way to evaluate the LLM is the accuracy, that is

\begin{equation}
    \mathrm{Acc}_{\mathrm{multi}} = \frac{\sum_{i = 1}^{L}y_{{ii}}}{\sum_{i = 1}^{L}{\sum_{j = 1}^{L}y_{{ij}}}}
\end{equation}

To overcome the multi-label confusion, both the macro-F1 score and micro-F1 score treat each label separately as a single classification problem. For example, for the instances which belong to the label \(i\), the label \(i\) is regarded as the positive, and all the other labels are regarded as the negative. In other words, the \(y_{{ii}}\) is the true positive, \(\sum_{j \neq i}^{}y_{{ij}}\) is the false positive, \(\sum_{j \neq i}^{}y_{{ji}}\) is the false negative. The rest instances belong to the true negative. The micro-F1 harmonically averages the micro-precision and micro-recall. The micro-precision is

\begin{equation}
\mathrm{micro\mbox{-}precision}=\frac{\mathrm{Total\ TP}}{\mathrm{Total\ TP + Total\ FP}}= \frac{\sum_{i = 1}^{L}y_{{ii}}}{\sum_{i = 1}^{L}\left( y_{{ii}} + \sum_{j \neq i}^{}y_{{ij}} \right)}
\end{equation}

While the micro-recall is

\begin{equation}
\mathrm{micro\mbox{-}recall}=\frac{\mathrm{Total\ TP}}{\mathrm{Total\ TP + Total\ FN}} = \frac{\sum_{i = 1}^{L}y_{{ii}}}{\sum_{i = 1}^{L}\left( y_{{ii}} + \sum_{j \neq i}^{}y_{{ji}} \right)}
\end{equation}

Then the micro-F1 is the harmonic mean of micro-recall and micro-precision, that is

\begin{equation}
    \mathrm{Micro-F1}= \frac{2 \times \mathrm{micro\mbox{-}recall} \times \mathrm{micro\mbox{-}precision}}{\mathrm{micro\mbox{-}recall} + \mathrm{micro\mbox{-}precision}}
\end{equation}

The macro-F1 evaluates the LLM in another way by averaging the class F1 for each label. For label \(i\), the class F1 can be calculated by

\begin{equation}
    \mathrm{F1}_{i} = \frac{2\left(\frac{y_{{ii}}}{\sum_{(j \neq i)}^{}y_{{ji}} + y_{{ii}}}\right)\left(\frac{y_{{ii}}}{\sum_{j \neq i}^{}y_{{ij}} + y_{{ii}}}\right)}{\left(\frac{y_{{ii}}}{\sum_{(j \neq i)}^{}y_{{ji}} + y_{{ii}}}\right) + \left(\frac{y_{{ii}}}{\sum_{j \neq i}^{}y_{{ij}} + y_{{ii}}}\right)}
\end{equation}

Then the macro-F1 is

\begin{equation}
    \mathrm{Macro\mbox{-}F1} = \frac{1}{L}\sum_{i = 1}^{L}{
    \mathrm{F1}_{i}}
\end{equation}

The micro-F1 gives equal weight to each instance, which means it leans to the class with more instances, while the macro-F1 gives equal weight to each class. 

From a statistical perspective, both micro-precision and micro-recall quantify the probability that predicted labels exactly match the true labels, symbolized as $P(X=Y)$. This is equivalent to the multi-label accuracy, implying that these metrics essentially measure the same aspect of model performance. The micro-F1 score, calculated as the harmonic mean of micro-precision and micro-recall, similarly reflects $P(X=Y)$. Consequently, micro-F1 does not offer a distinct advantage over accuracy in evaluating LLMs, as it essentially conveys the same information. This analysis presupposes that the true label space is the same with the classifier's label space, sharing the same set of labels, as a common assumption in LLM evaluations. Contrastingly, the macro-F1 score takes a different approach by averaging the F1 scores of each class with equal weight, as denoted by the equation:
\begin{equation}
    \frac{1}{L}\sum_{i=1}^L\frac{2}{1/P(X=i\mid Y=i)+1/P(Y=i\mid X=i)}.
\end{equation}
Although this formula doesn't directly translate to a simple probability expression, it addresses a critical limitation of accuracy metrics: their tendency to overlook classes with fewer samples. This feature makes macro-F1 a preferred metric for evaluating LLMs, as it ensures that all classes, regardless of size, are considered equally. The widespread adoption of macro-F1 in LLM evaluations can be attributed to its ease of implementation and its ability to facilitate comparisons with previously developed models. Despite these considerations, more comprehensive evaluation methods, such as the ROC surface, remain underutilized in LLM assessments. These metrics, which can offer a more nuanced understanding of model performance in multi-label problems, have yet to become standard in the evaluation of LLMs.

\subsection{Token-Similarity(TS) Metrics}

Token-Similarity metrics encompass metrics that gauge the similarity between the texts generated by LLMs and the reference texts. They are instrumental in assessing how well LLMs can produce a desired sequence of words given contextual information. LLMs frequently employ these metrics to evaluate the quality of generated texts and their impact on tasks such as machine translation and text summarization. Key metrics in this category include Perplexity, BLEU (Bilingual Evaluation Understudy), ROUGE (Recall-Oriented Understudy for Gisting Evaluation)-1 or 2, ROUGE-L, and BertScore, METEOR (Metric for Evaluation of Translation with Explicit Ordering). They primarily assess LLM performance at the token level.

To narrow our focus, we omit the tokenization process performed by LLMs and assume that the texts are already adequately tokenized. We proceed with the understanding that the LLM generates texts as follows: Given the context \(\left\{ X_{- t} \right\}_{t = T}^{0}\), where \(T\) represents the total length of the context, the LLM predicts a sequence of tokens after this contextual information. These contextual texts could be either the prompt or the original texts that need to be summarized in LLMs. Let the predicted sequence of tokens by the LLM be represented as \({\{ x_{i}\}}_{i = 1}^{N}\). To assess the quality of the generated texts, benchmark datasets provide a reference sequence of tokens \({\{ y_{j}\}}_{j = 1}^{M}\), which serves as the gold standard.

LLMs assign probabilities to each token in the vocabulary and select tokens based on specific criteria to compose the desired generated texts. We denote the probabilities assigned by the LLM as $\hat{P}(\cdot)$. Perplexity, introduced by \cite{brown1988statistical}, focuses on measuring the occurrence probability of the reference sequence $\{ y_{j} \}_{j = 1}^{M}$ according to the LLM, formulated as:

\begin{equation}
    \mathrm{Perplexity} = 2^{-\frac{1}{M}\sum_{j=1}^{M}{\log_{2}{\hat{P}\left( y_{j} \right)}}},
\end{equation}
where $\hat{P}(y_{j})$ represents the probability assigned by the LLM to the $j$-th token in the reference sequence.

From a statistical perspective, Perplexity is inversely proportional to the likelihood function $\prod_{j = 1}^{M}{\hat{P}(y_{j})}$, thus lower perplexity values indicate better performance by the LLM in predicting the data.

BLEU proposed by \cite{papineni2001method} evaluates the LLM based on \(n\)-grams. An \(n\)-gram is a re-grouping of token-level sequences to measure the co-occurrences of tokens. Given a sequence \(\left\{ x_{i} \right\}_{i = 1}^{N}\), an \(n\)-gram indicates subsequences of length \(n\) as, 

\[\left\{ \left( x_{i},\ x_{i + 1},\cdots,\ x_{i + n - 1} \right) \right\}_{i = 1}^{N - n + 1}\]. 

Two \(n\)-grams match if every element matches in the given order, i.e., \(x_{i} = y_{j},\ x_{i + 1} = y_{j + 1},\ \cdots,\ x_{i + n - 1} = y_{j + n - 1}\). Then, the Precision between the generated texts \(\left\{ x_{i} \right\}_{i = 1}^{N}\) and the reference texts \(\left\{ y_{j} \right\}_{j = 1}^{M}\) at the \(n\)-gram level is given by:

\begin{equation}
\begin{array}{cc}
    \mathrm{Precision}_{n} &= \frac{\mathrm{Number\ of\ matching\ n\mbox{-}grams}}{\mathrm{Total\ number\ of\ n\mbox{-}grams\ in\ generated\ text}} \\&= \frac{\mathrm{Number\ of\ matching\ n\mbox{-}grams}}{N - n + 1}
\end{array}
\end{equation}

Token-level matching is a special case, equivalent to 1-gram matching. BLEU introduces a Brevity Penalty to penalize cases where the generated text is too short and only partially matches the reference text, neglecting key information. The Brevity Penalty is:

\begin{equation}
    \mathrm{Brevity\ Penalty (BP)} = \min\left(1,\frac{\mathrm{Tokens\ in\ generated\ text}}{\mathrm{Tokens\ in\ reference\ text}}\right) = \min\left(1,\frac{N}{M}\right)
\end{equation}

The first type of BLEU employs the formula:

\begin{equation}
    \mathrm{BLEU} = \mathrm{BP} \times \exp{(\log{(\mathrm{Precision}_{1})})}
\end{equation}

Another type combines precision for \(n\)-grams with orders 1-4:

\begin{equation}
    \mathrm{BLEU} = \mathrm{BP} \times \exp{\left(\frac{1}{4}\sum_{n = 1}^{4}{\log{\mathrm{Precision}_{n}}}\right)}
\end{equation}

The BLEU also possesses its statistical meaning. We assume the event that a token in the corpus to appear in the reference text is denoted by $A_1$, and to appear in the generated text is denoted by $B_1$. Similarly, the events for the n-$gram$ appearing in two texts are denoted by $A_n$ and $B_n$. Then the Precision$_1$ within the BLEU in the first form denotes $P(A_1 \cap B_1 \mid B_1 )$ and the BLEU is proportional to it. The second form of BLEU is proportional to the $\prod_{i=1}^4 P(A_i\cap B_i \mid B_i)$. BLEU primarily emphasizes precision but overlooks evaluations of recall. LLMs may achieve high BLEU scores by capturing partial information from the reference texts, even if failing to predict the entirety of the reference texts accurately. This limitation makes BLEU less suitable for researchers who require LLMs to capture all relevant information from the context and accurately predict the reference texts.

Different from BLEU, ROUGE-n and ROUGE-L \citep{lin2004rouge} are \(n\)-gram level F1 scores. ROUGE-n can be calculated as:

\begin{equation}
    \begin{array}{rl}
    \mathrm{Precision\mbox{-}n} &= \frac{\mathrm{Number\ of\ matching\ n\mbox{-}grams}}{\mathrm{Total\ number\ of\ n\mbox{-}grams\ in\ generated\ text}} \\&= \frac{\mathrm{Number\ of\ matching\ n\mbox{-}grams}}{N - n + 1}\\
    \mathrm{Recall\mbox{-}n} &= \frac{\mathrm{Number\ of\ matching\ n\mbox{-}grams}}{M - n + 1}\\
    \mathrm{ROUGE\mbox{-}n} &= \frac{2 \times \mathrm{Precision\mbox{-}n} \times \mathrm{Recall\mbox{-}n}}{\mathrm{Precision\mbox{-}n} + \mathrm{Recall\mbox{-}n}}
    \end{array}
\end{equation}

The most common ROUGE-n metrics are ROUGE-1 and ROUGE-2. ROUGE-L extends ROUGE-n by focusing on finding the Longest Common Subsequence (LCS), denoted as nLCS. ROUGE-L is calculated as the harmonic mean of Precision and Recall:

\begin{equation}
    \mathrm{ROUGE\mbox{-}L} = \frac{2 \times \frac{\mathrm{nLCS}}{M} \times \frac{\mathrm{nLCS}}{N}}{\frac{\mathrm{nLCS}}{M} + \frac{\mathrm{nLCS}}{N}}.
\end{equation}

The ROUGE-n can be expressed as a realization for 

\[
\frac{2}{1/P(A_n\cap B_n \mid B_n) + 1/P(A_n\cap B_n \mid A_n)},
\]
while the ROUGE-L can be similarly defined.

METEOR proposed by \cite{banerjee2005meteor} is a metric based on ROUGE but aims to mitigate the effects of different word variants and synonymy issues. Initially, both the reference texts and the generated texts are tokenized. Then, METEOR applies stemming to reduce the words to their base form in both texts. The stemmed sequences are denoted as \(\left\{ x_{i}^{\prime} \right\}_{i = 1}^{N}\) and \(\left\{ y_{j}^{\prime} \right\}_{j = 1}^{M}\). ROUGE is then applied to the stemmed sequences of texts, with ROUGE-1 being the most commonly used metric for METEOR. Upon ROUGE, METEOR also introduces a penalty term to reward the LLM for generating sequences of tokens in the same order as they appeared in the reference texts. The penalty is calculated based on the number of chunks. Let \(\left\{ z_{k} \right\}_{k = 1}^{K}\) represent the matched tokens ordered by their appearances in the generated texts. If a sequence of these tokens appears adjacently in both the generated and reference texts, they are combined into a chunk. For example, if the reference texts are "It is a guide to action" and the generated texts are "It is a guide to directing the learners", then the matched tokens include "It", "is", "a", "guide", and "to". However, since the combination "It is a guide to" appears in both texts in the same order, it is considered a chunk. The penalty is calculated as:

\begin{equation}
    \mathrm{Penalty} = \left( \frac{\# \mathrm{chunks}}{\# \mathrm{matched\ tokens}} \right)^{3}, 
\end{equation}
where $\#$ denotes the number. In an extreme case where the generated texts exactly match the reference texts, the number of chunks is only 1. A lower penalty indicates a better match between the generated and reference texts. When there are no bi-grams or longer matches, the number of chunks is equal to the number of matched tokens, suggesting the Penalty increases to 1. METEOR combines ROUGE and the Penalty as follows:

\begin{equation}
    \mathrm{METEOR} = \mathrm{ROUGE\mbox{-}1}(1-\beta\mathrm{Penalty}),
\end{equation}
where $\beta$ is selected as 0.5 by \cite{banerjee2005meteor}. In the worst-case scenario where there are no bi-grams or longer matches, METEOR's performance can decrease to half of that of ROUGE-1. Conversely, when there are mostly longer matches, METEOR closely aligns with ROUGE-1. Unlike ROUGE, METEOR tackles issues related to synonymy and word variants, prioritizing longer matches between the generated and reference texts.

BertScore proposed by \cite{zhang2019bertscore} also measures token similarities but not based on the matching of n-grams; instead, it relies on BERT \citep{devlin2018bert} pre-trained embeddings. Initially, it loads the BERT embeddings. For each token in the generated texts \(\left\{ x_{i} \right\}_{i = 1}^{N}\) and the reference texts \(\left\{ y_{j} \right\}_{j = 1}^{M}\), the corresponding embeddings can be found within the BERT embeddings, each represented as fixed-length vectors denoted by \(\left\{ {\widehat{\mathbf{x}}}_{\mathbf{i}}^{\mathbf{e}} \right\}_{i = 1}^{N}\) and \(\left\{ {\widehat{\mathbf{y}}}_{\mathbf{j}}^{\mathbf{e}} \right\}_{j = 1}^{M}\). These sequences of vectors are then zipped to a single vector representing the two texts. The usual approach is to element-wise average:

\begin{equation}
    {\widehat{\mathbf{x}}}^{\mathbf{e}} = \frac{1}{N} \sum_{i = 1}^{N}{\widehat{\mathbf{x}}}_{\mathbf{i}}^{\mathbf{e}}, \quad
    {\widehat{\mathbf{y}}}^{\mathbf{e}} = \frac{1}{M} \sum_{j = 1}^{M}{\widehat{\mathbf{y}}}_{\mathbf{j}}^{\mathbf{e}}.
\end{equation}

BertScore then calculates the cosine similarity of these two vectors as a measure of the similarity between the reference texts and the generated texts:
\begin{equation}
    \mathrm{BertScore} = \frac{{\widehat{x}}^{e} \cdot {\widehat{y}}^{e}}{||{\widehat{x}}^{e}||\cdot   ||{\widehat{y}}^{e}||}.
\end{equation}

The cosine similarity measures the angle between two vectors projected onto a multi-dimensional space, regardless of their size. It ranges from -1 to 1, with a higher value indicating greater similarity between the generated and reference texts. It aids in gauging the semantic similarities among documents, thus widely used in NLP. Given that cosine similarity maps the similarity of two vectors to the space $[-1, 1]$, statistical methods designed to measure correlation with high-dimensional random vectors can be applied. These include Pearson's correlation, Canonical Correlation \citep{borga2001canonical}, Distance Correlation \citep{szekely2013distance}, and others.

METEOR and BertScore represent more complex evaluation metrics for LLMs compared to those previously discussed. Their complexity arises from the reliance on external linguistic information, making a straightforward statistical explanation challenging. Unlike simpler metrics, METEOR and BertScore effectively address synonymy issues, offering a significant advantage in evaluating LLMs for nuanced linguistic understanding. These metrics have thus become widely adopted in the assessment of LLMs. It is noteworthy that, with the exception of perplexity, all metrics discussed in this section favor higher scores as indicators of superior LLM performance.

\subsection{Question-Answering(QA) Metrics}

Different from general classification problems or text matching problems, the Question-Answering task is a special task that involves fuzzy matching issues, making the application of previous metrics not straightforward. The QA task requires the LLM to identify the answer to a specific question given a contextual passage. The answer for this question is usually located in the contextual passage, so the QA task is transformed to predict the starting position and the ending position in the contextual passage, assuming the middle part is the answer. This is a restricted double-task prediction problem. An LLM often predicts the starting position and ending position separately and excludes combinations that are unreasonable, such as when the predicted ending position is before the predicted starting position. Among the remaining combinations, the LLM ranks them by the overall prediction probability for the starting position and ending position. Suppose there are overall \(N\) questions, each with a contextual passage. For question \(i\), the LLM places predictions in order from the most likely one to the less likely one, denoted by \((s_{i1},\ e_{i1}),\ (s_{i2},e_{i2}),\ \cdots\), where $s, e$ denotes the starting position and ending position located in the original text respectively. Assume the gold standard answer is located in the passage with starting position and ending position as \((s_{ik_{i}},\ e_{ik_{i}})\), where \(k_{i}\) is the rank of it among all predictions given by the LLM. There are three most popular metrics for these QA tasks. Strict Accuracy (SaCC) \citep{tsatsaronis2015overview} is the proportion of completely correct predictions:
\begin{equation}
    \mathrm{SaCC} = \frac{\#\left\{ k_{i} = 1 \right\}}{N},
\end{equation}
here \(k_{i} = 1\) means the optimal prediction ranking first among all the predictions is exactly the gold standard answer.

Lenient Accuracy (LaCC) \citep{tsatsaronis2015overview} is a more relaxed metric that allows partially correct predictions:
\begin{equation}
    \mathrm{LaCC} = \frac{\#\left\{ k_i \leq 5 \right\}}{N}.
\end{equation}
With Lenient Accuracy, if the top-5 predictions contain the gold standard answer, it is regarded as correct predictions.

Mean Reciprocal Rank (MRR) \citep{tsatsaronis2015overview} comprehensively evaluates the LLM, considering not only the optimal predictions but also suboptimal predictions perfectly matching the gold standard answer. It is the average of the reciprocal ranks of the correct predictions:

\begin{equation}
\mathrm{MRR} = \frac{1}{N}\sum_{i = 1}^{N}\frac{1}{k_{i}}.
\end{equation}

All the three metrics are highly related to the rank statistics. Assume the rank statistics for the gold standard answer is $R(S)$, then we have $SaCC=P(R(S)=1)$, $LaCC=P(R(S)<=5)$, and $MRR=E(1/R(S))$, where $E(\cdot)$ denotes the expectation. We summarize the statistical interpretations for all above metrics in Table~\ref{tab:interpret}.

\begin{table}
    \centering
    \begin{tabular}{ll}
    \hline Metrics & Statistical interpretation \\
    \hline Accuracy & $P(X=Y)$ \\
    Recall & $P(X=1\mid Y=1)$ \\
    Precision & $P(Y=1\mid X=1)$ \\
    F1-score & $2/(1/P(X=1\mid Y=1) + 1/P(Y=1\mid X=1))$ \\
    Micro-F1 & $P(X=Y)$ \\
    Macro-F1 & $1/L\sum_{i=1}^L 2/({1/P(X=i\mid Y=i)+1/P(Y=i\mid X=i)})$ \\
    \hline Perplexity & $(\mathrm{Likelihood}(\{y_i\}_{i=1}^M))^{-1/M}$ \\
    BLEU & $P(A_1 \cap B_1 \mid B_1 )$ \\
    ROUGE-n & $2/({1/P(A_n\cap B_n \mid B_n) + 1/P(A_n\cap B_n \mid A_n)})$ \\
    ROUGE-L & $2/({1/P(A_{LCS}\cap B_{LCS} \mid B_{LCS}) + 1/P(A_{LCS}\cap B_{LCS} \mid A_{LCS})})$ \\
    METEOR & \\
    BertScore & \\
    \hline SaCC & $P(R(S)=1)$ \\
    LaCC & $P(R(S)\leq 5)$ \\
    MRR & $E(1/R(S))$ \\
    \hline
    \end{tabular}
    \caption{Statistical interpretations for each metric}
    \label{tab:interpret}
\end{table}

So far, we have classified the most popular metrics used to evaluate the LLMs into three categories: Multi-Classification, Token-Similarity, and Question-Answering metrics, and introduced the mathematical formulations and statistical implications for each metric. To facilitate researchers in applying these metrics, we list the existing packages or functions for these methods in Python in Table~\ref{tab:metrics}. While there may be other metrics used to measure the efficiency of LLMs, most of them are less popular compared to the above metrics, so we did not provide their mathematical formulations. In the next section, we will showcase their applications with recently published LLMs in biomedical and global fields.

\begin{table}
    \centering
    \begin{tabular}{lll}
    \hline Classification & Metrics & Repositories or Python functions \\
    \hline \multirow{6}{*}{ Multi-Classification } & Accuracy & sklearn.metrics.accuracy\_score \\
    & Recall & sklearn.metrics.recall\_score \\
    & Precision & sklearn.metrics.precision\_score \\
    & F1-score & sklearn.metrics.f1
    \_score \\
    & Micro-F1 & sklearn.metrics.f1\_score(average=\texttt{'micro'}) \\
    & Macro-F1 & sklearn.metrics.f1\_score(average=\texttt{'macro'}) \\
    \hline \multirow{6}{*}{ Token-Similarities } & Perplexity & nltk.perplexity \\
    & BLEU & nltk.translate.bleu\_score \\
    & ROUGE-n & \url{https://github.com/pltrdy/rouge} \\
    & ROUGE-L & \url{https://github.com/pltrdy/rouge} \\
    & METEOR & \makecell[l]{nltk.translate.meteor\_score;\\
    \url{https://github.com/mcjoshi/qgeval}}\\
    & BertScore & \url{https://github.com/Tiiiger/bert score} \\
    \hline \multirow{3}{*}{ Question-Answering } & SaCC & Manual Implementation \\
    & LaCC & Manual Implementation \\
    & MRR & Manual Implementation \\
    \hline
    \end{tabular}
    \caption{Repositories or Python functions to realized the metrics}
    \label{tab:metrics}
\end{table}

\section{Application of these metrics: examples with biomedical LLMs}

We highlight the use of metrics through recently developed biomedical LLMs, spanning a broad spectrum from specialized biomedical to general corpora. The advent of pre-trained biomedical LLMs enhances capabilities such as abstract summarization within biomedical texts, specific question answering, and task completion relevant to medical contexts, like associating treatments with symptoms or clinical histories. Our literature review spanned databases like Google Scholar, PubMed, ArXiV, and ACM digital libraries, focusing on works from 2018 onwards that significantly relate to biomedical LLMs using keywords like "Medical/Biomedical/Clinical/Radiology" and "Large Language Model/Pre-trained model". While not exhaustive, our selection is representative, focusing on widely referenced works. We delineated applications of these metrics in biomedical LLMs, compiling training and benchmark datasets, and outlined the tasks applicable to each LLM, aiding researchers in benchmarking their models against the previous competitive LLMs.

We provide a concise listing of biomedical LLMs in the Supplementary Information B due to the length of our paper, detailing their training datasets alongside the downstream tasks mentioned in the paper for each model, showcasing their performance capabilities. Before diving into these biomedical LLMs, we offer a concise overview of downstream tasks in LLMs, which play an important role in evaluating the performance of the LLMs. We left the overview of the downstream tasks in the Supplementary Information A. From the listing of existing biomedical LLMs, the PubMed and PMC corpus emerge as the most frequently utilized training datasets across the biomedical LLM landscape, reflecting their extensive adoption for model training. We also found that most LLMs showcase their capabilities across multiple downstream tasks but not a single one. 

We further illustrate the application of various metrics in evaluating these biomedical LLMs, as detailed in Table~\ref{tab:metrics_use}. A checkbox in the table indicates the utilization of a specific metric for evaluation in the corresponding LLM. Beyond the aforementioned metrics, additional metrics are occasionally employed in some studies, though they represent a minor fraction of the papers reviewed. Among them, MC metrics are predominantly favored across all three categories of metrics due to their straightforward applicability to established benchmark datasets. For TS metrics, ROUGE-n and ROUGE-L are the most commonly used, acting as F1-scores at the token level between reference and generated texts. QA metrics are often applied together. Despite being specifically designed for Question Answering tasks, there's a noticeable trend towards employing MC metrics with redesigned QA benchmarks as an alternative to QA-specific metrics. For instance, both BioInstruct \citep{bioinstruct} and RoBERTa \citep{roberta} include QA tasks in their evaluations but rely solely on MC metrics and BERTScore. This preference for simpler MC metrics facilitates alignment with other LLMs but might neglect the inherent challenges LLMs face in fully comprehending human language and ensuring the fluency and comprehensibility of generated text. Therefore, a comprehensive evaluation incorporating a broad spectrum of metrics is advisable for a thorough assessment of an LLM's capabilities.

\begin{table}
    \centering
    \begin{adjustbox}{width=\textwidth}
    \begin{tabular}{ccccccccccccc}
    \hline Biomedical LLMs & Acc & Recall & Precision & F1/macro-F1 & BLEU & ROUGE-1/2/L & METEOR & BERT-Score & MRR & SaCC& LaCC & Additional \\
    \hline BioBERT & & \checkmark & \checkmark & \checkmark & & & & & \checkmark & & & \\
    BioGPT & \checkmark & \checkmark & \checkmark & \checkmark & & & & & & & & \\
    BioLinkBERT & \checkmark & & & \checkmark & & & & & & & & \\
    BioMedGPT & \checkmark & & & & \checkmark & \checkmark & \checkmark & & & & & \\
    BioMegatron & & \checkmark & \checkmark & \checkmark & & & & & \checkmark & \checkmark & \checkmark & \\
    ClinicalBERT & \checkmark & \checkmark & \checkmark & \checkmark & & & & & & & & \\
    DoT5 & \checkmark & & & \checkmark & \checkmark & \checkmark & & & & & & \makecell[c]{NEM, \\cheXbert} \\
    ELECTRAMed & & \checkmark & \checkmark & \checkmark & & & & & \checkmark & \checkmark & \checkmark & \\
    GatorTronGPT & & \checkmark & \checkmark & \checkmark & & & & & & & & \makecell{Pearson-\\Correlation} \\
    MedPaLM & \checkmark &  &  & &  & & & &  & & & \\
    MedPaLM2 & \checkmark & & & & & & & & & & & \\
    PubMedBERT & & & & & & & & & & & & BLURB \\
    SciBERT & & & & \checkmark & & & & & & & & \\
    SciFive & & & & \checkmark & & & & & & & & \\
    GenCompareSum & & & & & & \checkmark & & & & & & \\
    RadBERT & \checkmark & & & \checkmark & & \checkmark & & & & & & \\
    BioBERTsum & & & & & & \checkmark & & & & & & \\
    BioBART & & \checkmark & & \checkmark & \checkmark & \checkmark & & \checkmark & & & & \\
    KeBioSum & & & & \checkmark & & \checkmark & & \checkmark & & & & \\
    Biolnstruct & & \checkmark & \checkmark & \checkmark & & & & \checkmark & & & & \\
    BioRoBERTa & & & & \checkmark & & & & & & & & \\
    RoBERTa & \checkmark & & & \checkmark & & & & & & & & \\
    BioELMo & \checkmark & & & & & & & & & & & \\
    
    \hline
    \end{tabular}
\end{adjustbox}
    \caption{The metrics used in the published paper for each LLM}
    \label{tab:metrics_use}
    
\end{table}

In addition to the metrics, we have also summarized the benchmark datasets employed in these biomedical LLMs, which is left in the Supplementary Information C. Among all the downstream tasks, NER, RE, QA and TS are the focal points for biomedical LLMs. Datasets such as BC5CDR, NCBI Disease, MedNLI, and CHEMPROT are frequently used, each by more than five LLMs, indicating their widespread adoption for evaluating model performance in these specific tasks.

\section{Strengths and weakness on the metrics}

The metrics in language model evaluations are categorized into Multiple-Classification (MC), Token-Similarity (TS), and Question-Answering (QA) types. MC metrics, primarily used in multiple classification tasks, are prevalent in assessing LLM performance. While MC metrics offer simplicity and alignment in LLM evaluations, they depend heavily on well-structured benchmark datasets with each subject assigned to a single label, assuming perfect labeling. This reliance on perfect labeling presents inherent limitations: firstly, creating such datasets demands significant resources; secondly, MC metrics struggle with subjects having ambiguous or multiple labels, necessitating more advanced metrics; finally, the assumption of perfect labeling overlooks the challenges of real-world data without human-assigned labels, potentially impacting LLM performance and generalization.

Moreover, an interesting parallel exists between MC metrics and the metrics used in diagnostic studies, particularly in their emphasis on sensitivity and specificity, along with ROC/AUC analysis. While diagnostic studies might prioritize sensitivity and specificity, LLM evaluations often focus on Recall and Precision, analogous to sensitivity and Positive Predictive Value (PPV), respectively. Challenges arise in the context of unbalanced datasets, where a classifier could misleadingly show high performance metrics by overpredicting the majority class. Supposing a dataset with only a few subjects labeled as negative, an extreme classifier predicting all the subjects as positive would obtain both high recall and precision, accordingly high F1-score. However, it is not a classifier that should be preferred. Thus it highlights the inadequacy of using conventional metrics like F1-score without considering the balance between classes or employing techniques like resampling or weighted scores. The aforementioned example of imbalanced data underscores the importance for researchers to meticulously select metrics tailored to the data structure. Additionally, for LLMs predicting only the label, a thorough evaluation with the pair of sensitivity/specificity and the pair of Recall/Prediction as well as the F1-score is recommended. For LLMs predicting labels with continuous biomarkers, methods that are free of the cut-off points can provide a more comprehensive assessment of model performance. When dealing with binary labels, consideration of AUC/PRAUC is essential, while for models with ordinal multiple labels, novel approaches such as ROC surface analysis are preferred. However, these methods remain underexplored in the context of existing LLMs. Additionally, the lack of an efficient selection threshold complicates the determination of LLM performance adequacy, as F1-score lacks robust statistical properties. Addressing these evaluation challenges necessitates the development of new metrics, possibly inspired by diagnostic studies, to ensure statistical reliability in LLM assessments. 

Compared to MC metrics, TS metrics assess the quality of generated texts by comparing them with original texts or aligning the provided answers accordingly. Among TS metrics, ROUGE-n and ROUGE-L \citep{lin2004rouge} are particularly prominent, effectively extending the F1-score to evaluate token-level similarity between reference and generated texts. However, these metrics assign equal importance to every token, not distinguishing between content-critical words (nouns, verbs, adjectives) and less impactful particles, potentially skewing the assessment of a text's semantic quality. Additionally, ROUGE metrics struggle with word variants and synonymy, making it challenging to capture the essence of longer texts comprehensively. Although METEOR \citep{banerjee2005meteor} and BERTScore \citep{zhang2019bertscore} aim to address these issues, they heavily rely on external techniques or pre-defined parameters. Exploring the development of more reliable metrics to evaluate token similarities and conducting a comprehensive evaluation with various metrics for LLMs would be intriguing.

QA metrics, tailored for Question Answering tasks, presume that answers reside within the original texts, necessitating meticulous benchmark dataset design. Traditional QA metrics, however, fall short in the era of abundant dialogue-based QA data, as they merely assess an LLM's ability to pinpoint answer boundaries within texts. Such metrics overlook ambiguously correct responses, where slight variations in answer positioning might still provide valid answers, underscoring the need for an automatic, versatile metric capable of handling free-form dialogue-type QA data without the constraints of traditional evaluation methods.

A major issue across all metric categories is imperfect labeling, known as an imperfect gold standard, which signifies inaccuracies or unreliability in reference labels or texts used for evaluation. Imperfect labeling manifests as misassigned labels in MC metrics, flawed reference texts in TS metrics, and ambiguously correct responses in QA metrics. It may stem from human error or the use of unverified techniques. While pre-trained LLMs like GPT-4 are increasingly employed to generate gold standards, they may introduce hallucinations, where LLMs may generate false or misleading information and the generated results are not aligned with user requests. Sensitivity analysis and statistical adjustment methods, primarily developed for diagnostic studies, offer solutions to mitigate bias arising from imperfect gold standards. \cite{umemneku2019diagnostic} conducted a systematic review of methods addressing imperfect gold standard bias in diagnostic studies. Studies by \cite{alonzo2005assessing} and \cite{to2016bias} employ imputation methods to mitigate bias, while others like \citep{brenner1996correcting,emerson2018biomarker,albert2009estimating} assume access to sensitivity and specificity information of the reference standard, leveraging it to adjust for bias. Yet, these methods are underutilized in LLM research. Overlooking such bias may lead researchers to draw incorrect conclusions. Borrowing ideas from correcting imperfect gold standard bias could offer new insights for evaluations in LLMs.

Another issue, particularly concerning TS metrics, is the absence of statistical inference methods. Most metrics solely provide a performance value for models without accompanying confidence intervals to gauge the reliability of the estimate. The intricate structures of texts may obscure the necessity of proposing statistical inference methods for these metrics. Nonetheless, this absence renders the metrics unreliable, as researchers cannot address the uncertainty surrounding the models' performance. A substantial variation may coincide with high LLM performance; however, such models may prove unsuitable for real-world applications.

\section{Conclusions}
\label{sec:conclusions}

Our study comprehensively reviews the most frequently utilized metrics in the evaluation of LLMs, showcasing their application through recently published biomedical LLMs. We aggregate benchmark datasets and downstream tasks associated with each LLM, marking the first comprehensive summary of metrics, benchmark datasets, and their evaluations in LLMs, complete with detailed mathematical formulations, statistical interpretations, and repositories for these metrics. This study aims to guide researchers in biomedical or other domain-specific/general LLMs in selecting appropriate benchmark datasets and evaluation metrics, facilitating better evaluation of their LLMs and comparison with competing models.

However, our study encounters several challenges. Firstly, while we encompass the most common evaluation metrics found in published literature, detailing every existing metric for all potential use cases is beyond our scope due to its complexity and the extensive manpower required. However, the metrics and recommendations we provide are representative and valuable. Secondly, we do not directly correlate benchmark datasets with specific metrics due to the flexibility in their application: a single dataset may be assessed using various metrics, and different LLMs might employ different metrics even when using the same dataset. Additionally, benchmark datasets may be adapted for particular tasks and metrics depending on the researchers' objectives. Thus, prescribing fixed metric usage for each benchmark dataset is neither feasible nor advisable. Nevertheless, a future study offering a comprehensive recommendation with detailed application scenarios would be of great interest.

\section*{Acknowledgments}
This research was funded by supported by National Key R\&D Program of China (No. 2021YFF0901400) and Novo Nordisk A/S.

\section*{Conflict of Interest}
The authors report there are no competing interests to declare.

\bigskip
\begin{center}
{\large\bf SUPPLEMENTARY MATERIAL}
\end{center}

\begin{description}

\item[Title:] Supplementary Information. (pdf)

\end{description}

\bibliography{JASA-template}

@article{bioinstruct,
  title={BioInstruct: Instruction Tuning of Large Language Models for Biomedical Natural Language Processing},
  author={Tran, Hieu and Yang, Zhichao and Yao, Zonghai and Yu, Hong},
  journal={arXiv preprint arXiv:2310.19975},
  year={2023}
}

@inproceedings{roberta,
  title={A robustly optimized BERT pre-training approach with post-training},
  author={Liu, Zhuang and Lin, Wayne and Shi, Ya and Zhao, Jun},
  booktitle={China National Conference on Chinese Computational Linguistics},
  pages={471--484},
  year={2021},
  organization={Springer}
}

@article{chen2023large,
  title={Large language models in biomedical natural language processing: benchmarks, baselines, and recommendations},
  author={Chen, Qingyu and Du, Jingcheng and Hu, Yan and Keloth, Vipina Kuttichi and Peng, Xueqing and Raja, Kalpana and Zhang, Rui and Lu, Zhiyong and Xu, Hua},
  journal={arXiv preprint arXiv:2305.16326},
  year={2023}
}

@article{tsatsaronis2015overview,
  title={An overview of the BIOASQ large-scale biomedical semantic indexing and question answering competition},
  author={Tsatsaronis, George and Balikas, Georgios and Malakasiotis, Prodromos and Partalas, Ioannis and Zschunke, Matthias and Alvers, Michael R and Weissenborn, Dirk and Krithara, Anastasia and Petridis, Sergios and Polychronopoulos, Dimitris and others},
  journal={BMC bioinformatics},
  volume={16},
  number={1},
  pages={1--28},
  year={2015},
  publisher={BioMed Central}
}

@article{sayers2023database,
  title={Database resources of the National Center for Biotechnology Information in 2023},
  author={Sayers, Eric W and Bolton, Evan E and Brister, J Rodney and Canese, Kathi and Chan, Jessica and Comeau, Donald C and Farrell, Catherine M and Feldgarden, Michael and Fine, Anna M and Funk, Kathryn and others},
  journal={Nucleic acids research},
  volume={51},
  number={D1},
  pages={D29--D38},
  year={2023},
  publisher={Oxford University Press}
}

@article{bengio2000neural,
  title={A neural probabilistic language model},
  author={Bengio, Yoshua and Ducharme, R{\'e}jean and Vincent, Pascal},
  journal={Advances in neural information processing systems},
  volume={13},
  year={2000}
}

@inproceedings{morin2005hierarchical,
  title={Hierarchical probabilistic neural network language model},
  author={Morin, Frederic and Bengio, Yoshua},
  booktitle={International workshop on artificial intelligence and statistics},
  pages={246--252},
  year={2005},
  organization={PMLR}
}

@article{weidinger2021ethical,
  title={Ethical and social risks of harm from language models},
  author={Weidinger, Laura and Mellor, John and Rauh, Maribeth and Griffin, Conor and Uesato, Jonathan and Huang, Po-Sen and Cheng, Myra and Glaese, Mia and Balle, Borja and Kasirzadeh, Atoosa and others},
  journal={arXiv preprint arXiv:2112.04359},
  year={2021}
}

@article{ganguli2022red,
  title={Red teaming language models to reduce harms: Methods, scaling behaviors, and lessons learned},
  author={Ganguli, Deep and Lovitt, Liane and Kernion, Jackson and Askell, Amanda and Bai, Yuntao and Kadavath, Saurav and Mann, Ben and Perez, Ethan and Schiefer, Nicholas and Ndousse, Kamal and others},
  journal={arXiv preprint arXiv:2209.07858},
  year={2022}
}

@article{jobin2019global,
  title={The global landscape of AI ethics guidelines},
  author={Jobin, Anna and Ienca, Marcello and Vayena, Effy},
  journal={Nature machine intelligence},
  volume={1},
  number={9},
  pages={389--399},
  year={2019},
  publisher={Nature Publishing Group UK London}
}

@article{bolukbasi2016man,
  title={Man is to computer programmer as woman is to homemaker? debiasing word embeddings},
  author={Bolukbasi, Tolga and Chang, Kai-Wei and Zou, James Y and Saligrama, Venkatesh and Kalai, Adam T},
  journal={Advances in neural information processing systems},
  volume={29},
  year={2016}
}

@inproceedings{dixon2018measuring,
  title={Measuring and mitigating unintended bias in text classification},
  author={Dixon, Lucas and Li, John and Sorensen, Jeffrey and Thain, Nithum and Vasserman, Lucy},
  booktitle={Proceedings of the 2018 AAAI/ACM Conference on AI, Ethics, and Society},
  pages={67--73},
  year={2018}
}

@inproceedings{hovy2021importance,
  title={The importance of modeling social factors of language: Theory and practice},
  author={Hovy, Dirk and Yang, Diyi},
  booktitle={Proceedings of the 2021 Conference of the North American Chapter of the Association for Computational Linguistics: Human Language Technologies},
  pages={588--602},
  year={2021}
}

@inproceedings{selbst2019fairness,
  title={Fairness and abstraction in sociotechnical systems},
  author={Selbst, Andrew D and Boyd, Danah and Friedler, Sorelle A and Venkatasubramanian, Suresh and Vertesi, Janet},
  booktitle={Proceedings of the conference on fairness, accountability, and transparency},
  pages={59--68},
  year={2019}
}

@article{raffel2020exploring,
  title={Exploring the limits of transfer learning with a unified text-to-text transformer},
  author={Raffel, Colin and Shazeer, Noam and Roberts, Adam and Lee, Katherine and Narang, Sharan and Matena, Michael and Zhou, Yanqi and Li, Wei and Liu, Peter J},
  journal={The Journal of Machine Learning Research},
  volume={21},
  number={1},
  pages={5485--5551},
  year={2020},
  publisher={JMLRORG}
}

@article{hupkes2023taxonomy,
  title={A taxonomy and review of generalization research in NLP},
  author={Hupkes, Dieuwke and Giulianelli, Mario and Dankers, Verna and Artetxe, Mikel and Elazar, Yanai and Pimentel, Tiago and Christodoulopoulos, Christos and Lasri, Karim and Saphra, Naomi and Sinclair, Arabella and others},
  journal={Nature Machine Intelligence},
  volume={5},
  number={10},
  pages={1161--1174},
  year={2023},
  publisher={Nature Publishing Group UK London}
}

@article{lazaridou2021mind,
  title={Mind the gap: Assessing temporal generalization in neural language models},
  author={Lazaridou, Angeliki and Kuncoro, Adhi and Gribovskaya, Elena and Agrawal, Devang and Liska, Adam and Terzi, Tayfun and Gimenez, Mai and de Masson d'Autume, Cyprien and Kocisky, Tomas and Ruder, Sebastian and others},
  journal={Advances in Neural Information Processing Systems},
  volume={34},
  pages={29348--29363},
  year={2021}
}

@article{wang2021measure,
  title={Measure and improve robustness in nlp models: A survey},
  author={Wang, Xuezhi and Wang, Haohan and Yang, Diyi},
  journal={arXiv preprint arXiv:2112.08313},
  year={2021}
}

@article{goel2021robustness,
  title={Robustness gym: Unifying the NLP evaluation landscape},
  author={Goel, Karan and Rajani, Nazneen and Vig, Jesse and Tan, Samson and Wu, Jason and Zheng, Stephan and Xiong, Caiming and Bansal, Mohit and R{\'e}, Christopher},
  journal={arXiv preprint arXiv:2101.04840},
  year={2021}
}

@article{goyal2023survey,
  title={A survey of adversarial defenses and robustness in nlp},
  author={Goyal, Shreya and Doddapaneni, Sumanth and Khapra, Mitesh M and Ravindran, Balaraman},
  journal={ACM Computing Surveys},
  volume={55},
  number={14s},
  pages={1--39},
  year={2023},
  publisher={ACM New York, NY}
}

@article{borga2001canonical,
  title={Canonical correlation: a tutorial},
  author={Borga, Magnus},
  journal={On line tutorial http://people. imt. liu. se/magnus/cca},
  volume={4},
  number={5},
  year={2001}
}

@article{szekely2013distance,
  title={The distance correlation t-test of independence in high dimension},
  author={Sz{\'e}kely, G{\'a}bor J and Rizzo, Maria L},
  journal={Journal of Multivariate Analysis},
  volume={117},
  pages={193--213},
  year={2013},
  publisher={Elsevier}
}

@article{umemneku2019diagnostic,
  title={Diagnostic test evaluation methodology: a systematic review of methods employed to evaluate diagnostic tests in the absence of gold standard--an update},
  author={Umemneku Chikere, Chinyereugo M and Wilson, Kevin and Graziadio, Sara and Vale, Luke and Allen, A Joy},
  journal={PLoS One},
  volume={14},
  number={10},
  pages={e0223832},
  year={2019},
  publisher={Public Library of Science San Francisco, CA USA}
}

@article{alonzo2005assessing,
  title={Assessing accuracy of a continuous screening test in the presence of verification bias},
  author={Alonzo, Todd A and Pepe, Margaret Sullivan},
  journal={Journal of the Royal Statistical Society Series C: Applied Statistics},
  volume={54},
  number={1},
  pages={173--190},
  year={2005},
  publisher={Oxford University Press}
}

@article{to2016bias,
  title={Bias--corrected methods for estimating the receiver operating characteristic surface of continuous diagnostic tests},
  author={To Duc, Khanh and Chiogna, Monica and Adimari, Gianfranco},
  year={2016}
}

@article{brenner1996correcting,
  title={Correcting for exposure misclassification using an alloyed gold standard},
  author={Brenner, Hermann},
  journal={Epidemiology},
  pages={406--410},
  year={1996},
  publisher={JSTOR}
}

@article{emerson2018biomarker,
  title={Biomarker validation with an imperfect reference: Issues and bounds},
  author={Emerson, Sarah C and Waikar, Sushrut S and Fuentes, Claudio and Bonventre, Joseph V and Betensky, Rebecca A},
  journal={Statistical methods in medical research},
  volume={27},
  number={10},
  pages={2933--2945},
  year={2018},
  publisher={SAGE Publications Sage UK: London, England}
}

@article{albert2009estimating,
  title={Estimating diagnostic accuracy of multiple binary tests with an imperfect reference standard},
  author={Albert, Paul S},
  journal={Statistics in medicine},
  volume={28},
  number={5},
  pages={780--797},
  year={2009},
  publisher={Wiley Online Library}
}

@article{saito2015precision,
  title={The precision-recall plot is more informative than the ROC plot when evaluating binary classifiers on imbalanced datasets},
  author={Saito, Takaya and Rehmsmeier, Marc},
  journal={PloS one},
  volume={10},
  number={3},
  pages={e0118432},
  year={2015},
  publisher={Public Library of Science San Francisco, CA USA}
}

@inproceedings{ghamrawi2005collective,
  title={Collective multi-label classification},
  author={Ghamrawi, Nadia and McCallum, Andrew},
  booktitle={Proceedings of the 14th ACM international conference on Information and knowledge management},
  pages={195--200},
  year={2005}
}

@inproceedings{goutte2005probabilistic,
  title={A probabilistic interpretation of precision, recall and F-score, with implication for evaluation},
  author={Goutte, Cyril and Gaussier, Eric},
  booktitle={European conference on information retrieval},
  pages={345--359},
  year={2005},
  organization={Springer}
}

@article{devlin2018bert,
  title={Bert: Pre-training of deep bidirectional transformers for language understanding},
  author={Devlin, Jacob and Chang, Ming-Wei and Lee, Kenton and Toutanova, Kristina},
  journal={arXiv preprint arXiv:1810.04805},
  year={2018}
}

@article{zhang2019bertscore,
  title={Bertscore: Evaluating text generation with bert},
  author={Zhang, Tianyi and Kishore, Varsha and Wu, Felix and Weinberger, Kilian Q and Artzi, Yoav},
  journal={arXiv preprint arXiv:1904.09675},
  year={2019}
}

@article{valmeekam2022large,
  title={Large Language Models Still Can't Plan (A Benchmark for LLMs on Planning and Reasoning about Change)},
  author={Valmeekam, Karthik and Olmo, Alberto and Sreedharan, Sarath and Kambhampati, Subbarao},
  journal={arXiv preprint arXiv:2206.10498},
  year={2022}
}

@inproceedings{jin2023cladder,
  title={Cladder: Assessing causal reasoning in language models},
  author={Jin, Zhijing and Chen, Yuen and Leeb, Felix and Gresele, Luigi and Kamal, Ojasv and Zhiheng, LYU and Blin, Kevin and Adauto, Fernando Gonzalez and Kleiman-Weiner, Max and Sachan, Mrinmaya and others},
  booktitle={Thirty-seventh Conference on Neural Information Processing Systems},
  year={2023}
}

@article{wang2022self,
  title={Self-instruct: Aligning language model with self generated instructions},
  author={Wang, Yizhong and Kordi, Yeganeh and Mishra, Swaroop and Liu, Alisa and Smith, Noah A and Khashabi, Daniel and Hajishirzi, Hannaneh},
  journal={arXiv preprint arXiv:2212.10560},
  year={2022}
}

@inproceedings{lin2004rouge,
  title={Rouge: A package for automatic evaluation of summaries},
  author={Lin, Chin-Yew},
  booktitle={Text summarization branches out},
  pages={74--81},
  year={2004}
}

@inproceedings{banerjee2005meteor,
  title={METEOR: An automatic metric for MT evaluation with improved correlation with human judgments},
  author={Banerjee, Satanjeev and Lavie, Alon},
  booktitle={Proceedings of the acl workshop on intrinsic and extrinsic evaluation measures for machine translation and/or summarization},
  pages={65--72},
  year={2005}
}

@article{papineni2001method,
  title={A method for automatic evaluation of machine translation''},
  author={Papineni, Kishore and Roukos, Salim and Ward, Todd and Zhu, W``BLEU},
  journal={the Proceedings of ACL-2002, ACL, Philadelphia, PA, July 2002},
  year={2001}
}

@inproceedings{brown1988statistical,
  title={A statistical approach to language translation},
  author={Brown, Peter F and Cocke, John and Della Pietra, Stephen A and Della Pietra, Vincent J and Jelinek, Frederick and Mercer, Robert L and Roossin, Paul},
  booktitle={Coling Budapest 1988 Volume 1: International Conference on Computational Linguistics},
  year={1988}
}

@article{wang2023pre,
  title={Pre-trained language models in biomedical domain: A systematic survey},
  author={Wang, Benyou and Xie, Qianqian and Pei, Jiahuan and Chen, Zhihong and Tiwari, Prayag and Li, Zhao and Fu, Jie},
  journal={ACM Computing Surveys},
  volume={56},
  number={3},
  pages={1--52},
  year={2023},
  publisher={ACM New York, NY}
}

@article{brown2020language,
  title={Language models are few-shot learners},
  author={Brown, Tom and Mann, Benjamin and Ryder, Nick and Subbiah, Melanie and Kaplan, Jared D and Dhariwal, Prafulla and Neelakantan, Arvind and Shyam, Pranav and Sastry, Girish and Askell, Amanda and others},
  journal={Advances in neural information processing systems},
  volume={33},
  pages={1877--1901},
  year={2020}
}

@article{chang2023survey,
  title={A survey on evaluation of large language models},
  author={Chang, Yupeng and Wang, Xu and Wang, Jindong and Wu, Yuan and Zhu, Kaijie and Chen, Hao and Yang, Linyi and Yi, Xiaoyuan and Wang, Cunxiang and Wang, Yidong and others},
  journal={arXiv preprint arXiv:2307.03109},
  year={2023}
}

@article{naveed2023comprehensive,
  title={A comprehensive overview of large language models},
  author={Naveed, Humza and Khan, Asad Ullah and Qiu, Shi and Saqib, Muhammad and Anwar, Saeed and Usman, Muhammad and Barnes, Nick and Mian, Ajmal},
  journal={arXiv preprint arXiv:2307.06435},
  year={2023}
}
\end{document}